\documentclass[preprint,journal]{vgtc}                

\ifpdf
  \pdfoutput=1\relax                   
  \usepackage{graphicx}                
  \DeclareGraphicsExtensions{.pdf,.png,.jpg,.jpeg} 
\else
  \usepackage{graphicx}                
  \DeclareGraphicsExtensions{.eps}     
\fi%

\graphicspath{{figures/}{pictures/}{images/}{./}{figs/}} 

\usepackage{booktabs}
\usepackage{microtype}                 
\PassOptionsToPackage{warn}{textcomp}  
\usepackage{textcomp}                  
\usepackage{mathptmx}                  
\usepackage{times}                     
\usepackage{cite}
\usepackage{amsfonts}
\usepackage[]{algorithm2e}
\usepackage[]{amsmath}
\onlineid{0}

\usepackage{xcolor}

\title{LSTMVis: A Tool for Visual Analysis of Hidden State Dynamics in Recurrent Neural Networks }

\author{Hendrik Strobelt, Sebastian Gehrmann, Hanspeter Pfister, and Alexander M. Rush \\ -- Harvard School of Engineering and Applied Sciences --   }
 \authorfooter{
 \item HS -- contact: hendrik@strobelt.com
 \item SG, HP, AR -- contact: \{gehrmann, pfister,rush\}@seas.harvard.edu. }
\shortauthortitle{Biv \MakeLowercase{\textit{et al.}}: Global Illumination for Fun and Profit}

\abstract{ Recurrent neural networks, and in particular long short-term
  memory (LSTM) networks, are a remarkably effective tool for
  sequence modeling that learn a dense black-box hidden
  representation of their sequential input. Researchers interested in
  better understanding these models have studied the changes in hidden
  state representations over time and noticed some interpretable
  patterns but also significant noise.  In this work, we present
  \textsc{LSTMVis}, a visual analysis tool for recurrent neural
  networks with a focus on understanding these hidden state dynamics.
  The tool allows users to select a hypothesis input range to focus
  on local state changes, to match these states changes to similar
  patterns in a large data set, and to align these results with
  structural annotations from their domain. We show several
  use cases of the tool for analyzing specific hidden state properties
  on dataset containing nesting, phrase structure, and chord
  progressions, and demonstrate how the tool can be used to isolate 
  patterns for further statistical analysis. We characterize the domain, the different
  stakeholders, and their goals and tasks. Long-term usage data after putting 
  the tool online revealed great interest in the machine learning community.
 
} 

\teaser{
   \centering
   \includegraphics[]{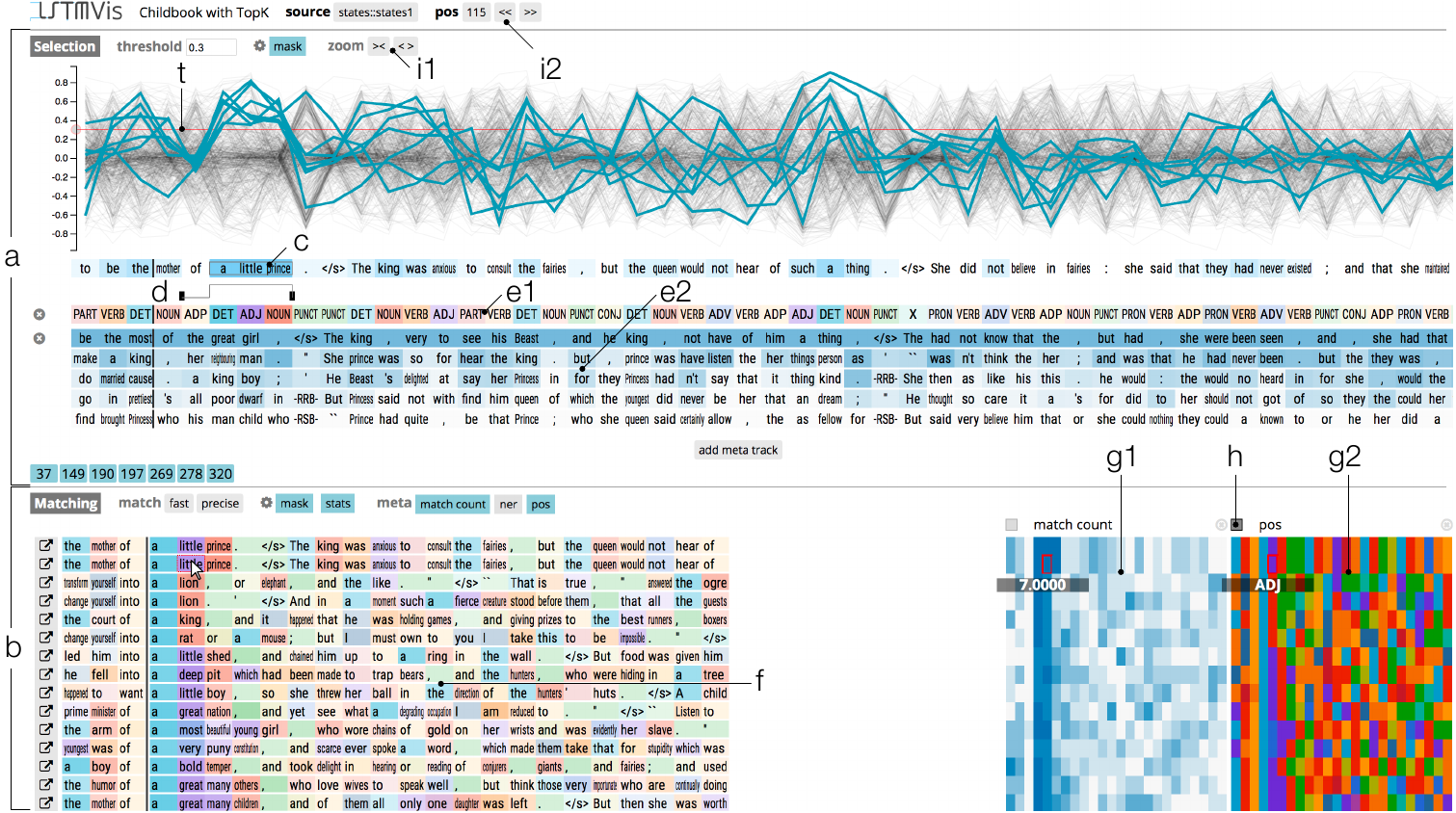}
   \caption{The \textsc{LSTMVis} user interface. The user
     interactively \textit{selects} a range of text specifying a hypothesis about the model in the Select View (a). This range is then used to \textit{match} similar hidden state patterns displayed in the Match View (b). The selection is made by specifying a start-stop range in the text (c) and an activation threshold (t) which leads to a selection of hidden states (blue lines). The start-stop range can be further constrained using the pattern plot (d). The meta-tracks below depict extra information per word position like POS (e1) or the top K predictions (e2). The tool can then \textit{match} this selection with similar hidden state patterns in the data set of varying lengths (f), providing insight into the representations learned by the model. The match view additionally includes user-defined meta-data encoded as heatmaps (g1,g2). The color of one heatmap (g2) can be mapped (h) to the word matrix (f) which allows the user to see patterns that lead to further refinement of the selection hypothesis. Navigation aids provide convenience (i1, i2).      }
\label{fig:teaser}
}

\vgtcinsertpkg

\begin{document}

\firstsection{Introduction}

\maketitle

In recent years, deep neural networks have become a central modeling tool for many artificial cognition tasks, such as image recognition, speech recognition, and text classification. These models all share a common property in that they utilize a \textit{hidden} feature representation of their input, not pre-specified by the user, which is learned for the task at hand. These hidden representations have proven to be very effective for classification. However, 
the black-box nature of these learned representations make the models themselves difficult to interpret. So while it is possible for users to produce high-performing systems, it is difficult for them to analyze what the system has learned. 

While all deep neural networks utilize hidden features, different model structures have shown to be effective for different tasks. Standard deep neural networks (DNNs) 
learn fixed-size features, whereas convolutional neural networks (CNNs), dominant in image recognition, will learn a task-specific filter-bank to produce spatial feature maps. In this work, we focus on deep neural network architectures known as recurrent neural networks (RNNs) that produce
a time-series of hidden feature-state representations.

RNNs~\cite{elman1990finding} have proven to be an effective general-purpose approach for capturing representation in sequence-modeling applications, such as text processing. Recent strong empirical results indicate that internal representations learn to capture complex relationships between the words within a sentence or document. These improved representation have led directly to end applications in machine translation~\cite{kalchbrenner2013recurrent,sutskever2014sequence}, speech recognition~\cite{DBLP:journals/corr/AmodeiABCCCCCCD15}, music generation~\cite{DBLP:conf/icml/Boulanger-LewandowskiBV12}, and text classification~\cite{dai2015semi}, among a variety of other applications.

While RNNs have shown clear improvements for sequence modeling, it has proven very difficult to interpret their feature representation. As such, it remains unclear exactly
\textit{how} a particular model is representing long-distance relationships within a sequence. Typically, RNNs contain millions of parameters and utilize repeated transformations of large hidden representations under time-varying conditions. These factors make the model inter-dependencies challenging to interpret without sophisticated mathematical tools. How do we enable users to explore complex network interactions in an RNN and directly connect these abstract representations to human understandable inputs?

In this work, we focus on the visual analysis of hidden features in RNNs. We have developed \textsc{LSTMVis}, a tool 
to allow advanced user groups to explore and form hypotheses about RNN hidden state dynamics. We analyzed neural network users and identified three major user roles, each with a different set of requirements: \textit{architects} who develop novel deep learning structures, \textit{trainers} who develop new data sets to train existing models, and \textit{end users} who apply deep models to new data. LSTMVis is focused on architects and trainers, for which we performed a goal and task analysis to develop effective visual encodings and interactions. LSTMVis combines a time-series based \textit{select} interface with an interactive \textit{match} tool to search for similar hidden state patterns in a large dataset. A live system can be accessed via \url{lstm.seas.harvard.edu} and the source code is provided. We present use cases applying our technique to identify and explore patterns in RNNs trained on large real-world datasets for text, speech recognition, biological sequence analysis and other domains. We discuss our release-before-publish strategy used to develop and improve the tool based on user feedback.

\section{Background: Recurrent Neural Networks}
\label{sec:back}

\begin{figure}
\centering
\includegraphics[width=\linewidth]{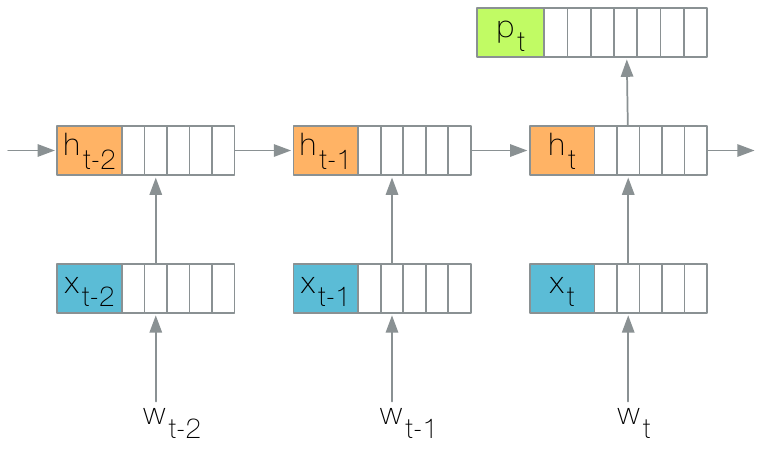}
\caption{ A recurrent neural network  language model being used
  to compute $p(w_{t+1} | w_1, \ldots, w_{t})$. At each time step, a word $w_t$ is converted to a word vector $\mathbf{x}_t$, which is then used to update the hidden state $\mathbf{h_t} \gets \mathbf{RNN}(\mathbf{x}_t, \mathbf{h}_{t-1})$. This hidden state vector can be used for prediction. In language modeling (shown) it is used to define the probability of the next word, $p(w_{t+1} | w_1, \ldots, w_{t}) = \text{softmax}(\mathbf{W} \mathbf{h_t} + \mathbf{b})$.
}
\label{fig:RNN}
\end{figure}

RNNs are a type of deep neural network architecture that has proven effective for sequence modeling tasks such as text processing. A major challenge of working with variable-length text sequences is producing features that capture or summarize long-distance relations in the text. These relationships are particularly important for tasks that require processing and generating sequences such as machine
translation. RNN-based models effectively learn hidden representations at each time-step which are then used for decision making. We refer to the change in these representations over time as the \textit{hidden state dynamics} produced by the model. 

Throughout this work, we will assume that we are given a sequence of words $w_1, \ldots, w_T$ for time 1 to $T$. These might consist of English words that we want to translate or a sentence whose sentiment we would like to detect, or even some other symbolic input such as musical notes or code.  Additionally, we will assume that we have a mapping from each word into vector representation $\mathbf{x}_1, \ldots, \mathbf{x}_T$. This representation can either be a standard fixed mapping, such as word2vec~\cite{mikolov2013distributed}, or can be learned with the rest of the model.

Formally, RNNs are a class of neural networks that sequentially map input word vectors $\mathbf{x}_1 \ldots \mathbf{x}_T$ to a sequence of hidden feature-state representations $\mathbf{h}_1, \ldots, \mathbf{h}_T$. This is achieved  by learning the weights of a neural network $\mathbf{RNN}$, which is applied recursively at each time-step $t \in {1\ldots T}$:

\[ \mathbf{h_{t}}\gets \mathbf{RNN}(\mathbf{x_{t}}, \mathbf{h_{t-1}}) \] 

\noindent This function takes input vector $\mathbf{x_{t}}$ and a hidden state vector $\mathbf{\mathbf{h}_{t-1}}$ and gives a new hidden state vector $\mathbf{h}_t$.  Each hidden state vector $\mathbf{h}_t$ is in $\mathbb{R}^D$. These vectors, and particularly how they change over time, will be the main focus of this work. We  are interested in each $c \in \{1 \ldots D\}$ and in particular the change of a single
\textit{hidden state} $h_{t,c}$ as $t$ varies.

The model learns these hidden states to represent the features of the input words. As such, they can be learned for any modeling tasks utilizing discrete sequential input. One notable application is RNN language modeling~\cite{mikolov2010recurrent,Zaremba2014}, a core task in natural language processing. In language modeling, at time $t$ the prefix of words $w_1, \ldots, w_t$ is taken as input and the goal is to model the distribution over the next word $p(w_{t+1} | w_1, \ldots, w_t)$. An RNN is used to produce this distribution by applying multi-class classification
based on hidden feature-state vector $\mathbf{h}_{t}$. Formally we define this as $p(w_{t+1} | w_1, \ldots, w_{t}) = \text{softmax}(\mathbf{W} \mathbf{h_t} + \mathbf{b})$,
where $\mathbf{W}, \mathbf{b}$ are parameters. The full computation of an RNN language model is shown in Figure~\ref{fig:RNN}.

It has been widely observed that the hidden states are able to capture important information about the structure of the input sentence necessary to perform this prediction. However, it has been difficult to trace how this is captured and what exactly is learned. For instance, it has been shown that RNNs can count parentheses or match quotes, but is unclear whether RNNs naturally discover aspects of language such as phrases, grammar, or topics. In this work, we focus particularly on exploring this question by examining the dynamics of the hidden states through time.

\begin{figure*}[htbp]
	\center
	\includegraphics[width=\textwidth]{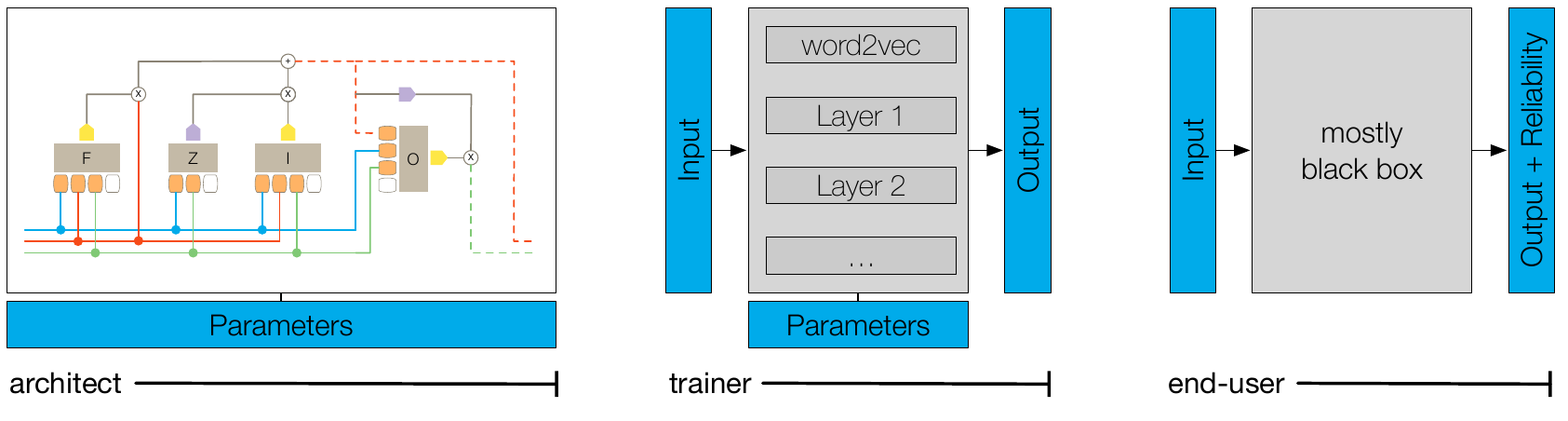}
	\caption{Views on neural network models for different user roles. The \textbf{architect} analyzes and modifies all components of the system. The \textbf{trainer} abstracts the model to the main components and parameters and focuses on training on different data sets. The \textbf{end user} has the most abstract view on the model and considers whether the output is coherent for a given input. }
	\label{fig:user_views}
\end{figure*}

Our use cases will mainly focus on long
short-term memory networks (LSTM)~\cite{hochreiter1997long}. 
LSTMs define a variant of RNN that has a modified hidden state update which can more effectively learn long-term interactions. As such these models are widely used in practice. In addition, LSTMs and RNNs can be \textit{stacked} in layers to produce multiple hidden state vectors at each time step, which further improves performance. While our results mainly use stacked LSTMs, our visualization only requires access to some time evolving abstract vector representation, and therefore can be used for
any layer of a wide variety of models.

Note that LSTMs maintain both a cell state vector and a hidden state vector at each time step. Our system can be used to analyze either or both of these vectors (or even the LSTM gates), and in our experiments we found that the cell states are easier to work with. For simplicity, however, we refer to these vectors generically as \textit{hidden states} throughout the paper.

\section{Related Work}
\label{sec:relwork}

\paragraph{Understanding RNNs through Visualization}
Our core contribution, visualizing the state dynamics of RNNs in a structured way, is inspired by previous work on convolutional neural networks in vision applications~\cite{simonyan2013deep,zeiler2014visualizing}. In linguistic tasks, visualizations have shown to be useful tool for understanding certain aspects of RNNs. Karpathy et al.~\cite{karpathy2015visualizing} use static visualization techniques to help understand hidden states in language models. Their work demonstrates that selected cells
can model clear events such as open parentheses and the start of URLs. Li et al.~\cite{li2015visualizing} present additional techniques, particularly the use of gradient-based saliency to find important words. Their work also looks at several different models and datasets including text classification and auto-encoders. Kadar et al.~\cite{kadar2016representation,kadar2015lingusitic} show
that RNNs specifically learn lexical categories and grammatical functions that carry semantic information, partially by modifying the inputs fed to the model. While inspired by these techniques, our approach tries to extend beyond single examples and provide a general interactive visualization approach of the raw data for exploratory
analysis.

\paragraph{Extending RNN Models for Interpretability} 
Recent work has also developed methods for extending RNNs for certain problems to make them easier to interpret (along with improving the models). One popular technique has been to use a neural \textit{attention} mechanism to allow the model to focus in on a particular aspect of the input. Bahdanau et al.~\cite{bahdanau2014neural} use attention for soft alignment in machine translation. Xu et al.~\cite{DBLP:conf/icml/XuBKCCSZB15} use attention to identify important aspects of an image for captioning, whereas Hermann et al.~\cite{DBLP:conf/nips/HermannKGEKSB15} use attention to find important aspects of a document for an extraction task. These approaches have the side benefit that they visualize the aspect of the model they are using. This approach differs from our work in that it requires changing the underlying model structure, whereas we attempt
to interpret the hidden states of a fixed model directly. 

\paragraph{Interactive Visualization of Neural Networks}
There has been some work on interactive visualization for interpreting machine learning models. Tzeng et al.~\cite{tzeng2005opening} present a visualization system for feed-forward neural networks with the goal of interpretation, and Kapoor et al.~\cite{kapoor2010interactive} give a user-interface for tuning the learning itself. The Prospector system~\cite{krause2016interacting} provides a general-purpose tool for practitioners to better understand their
machine learning model and its predictions. 

Recent work also describes systems that focus on analysis of hidden states for convolutional neural networks. Liu et al.~\cite{liu2017towards} utilize a DAG metaphor to show neurons, their connections, and learned features. Rauber et al.~\cite{rauber2017} use projections to explore relationships between neurons and learned observations. Other work has focused on user interfaces for constructing models, such as TensorBoard~\cite{abadi2016tensorflow} and the related playground for convolutional neural models at \url{playground.tensorflow.org/}. Our work is most similar in spirit to the work by Tzeng et al.~\cite{tzeng2005opening}, Liu et al.~\cite{liu2017towards}, and Rauber et al.~\cite{rauber2017}, in that we are concerned with interpreting the hidden states of neural network models. However, our specific goals focus on RNNs and the needs of specific users, and the resulting visual design is significantly different.

\section{User Analysis and Goals}
\label{sec:users}

Deep neural networks are now widely employed both in the research and industrial setting by a diverse set of users with different needs. Before developing our visual analysis goals we first laid out a set of prototypical stakeholders who might benefit from improved analysis. In meetings with members of a natural language processing group and a computational biology group, we identified three user roles, their incentives, and their view on neural network  models. Figure~\ref{fig:user_views} summarizes the following roles and which aspect of a model they 
might hope to master:

\begin{itemize}
\item \textbf{Architects} are looking to develop new deep learning methodologies or to modify existing deep architectures for new domains. An architect is interested
in training many variant network structures and comparing how the models capture the features of their domain. We assume that the architects are deeply knowledgeable about machine learning, neural networks, and the internal structure of the system. Their direct goal is comparing the performance of variant models and understanding the learned properties of the system. 

\item \textbf{Trainers} are those users interested in applying known architectures to tasks for which they are a domain experts. Trainers utilize RNNs as a tool and understand the key concepts of network optimization. However, their main focus is on the application domain and utilizing effective methods to solve known problems. Their goal is to use a known network architecture and to observe how it learns 
a novel model. Examples of trainers include a bioinformatician or an applied machine learning engineer.

\item \textbf{End Users} make up the most prevalent role of network users. End users utilize general-purpose pretrained networks for various tasks. These users may not need to understand the training process at all, only how to apply the networks as an algorithm to new data. Their main desire is to explain the results and locate what is happening when something goes wrong. Examples of end-users include data scientists or product engineers using ML.
\end{itemize}

\noindent
These user roles are general to the neural network domain and we believe they can help to describe and understand stakeholders in this space. For our analysis, we decided to focus  on the more advanced end of this spectrum, particularly on the user role of \textit{architects}. 
We aimed to provide these users with greater visibility into the internals of the system. User feedback from our first prototype further motivated us to include the \textit{trainer} role as users with a focus on the predictions of the model. In future work, we want to develop systems to engage the end users of RNN and other deep neural networks more.

With these roles in mind, we aimed to help trainers and architects better understand the high-level question: ``What information does an RNN capture in its hidden feature-states?''. Addressing this question is the main goal of our project. Based on a series of dicussions with deep learning experts we identified the following domain goals:

\begin{itemize}
\item \textbf{G1 - Formulate a hypothesis} about properties that the hidden states might learn to capture for a specific model. This hypothesis requires an initial understanding of hidden state values over time and a close read of the original input.
\item \textbf{G2 - Refine the hypothesis} based on insights about learned textual similarities based on patterns in the dynamics of the hidden states. Refining a hypothesis may also mean rejecting it.
\item \textbf{G3 - Compare models and datasets} to allow early generalization about the insights the representations provide, and to observe how task and domain may alter the patterns in the hidden states.
\end{itemize}

\noindent
During the design phase, we developed the following list of tasks for visual data analysis from the three domain goals (G1--G3). The mapping of these tasks to goals is indicated by square brackets:
\begin{itemize}
\item \textbf{T1 - Visualize hidden states} over time to allow exploration of the hidden state dynamics in their raw form. [G1]
\item \textbf{T2 - Filter hidden states} by using discrete textual selection along with continuous thresholding. These selections methods allow the user to form hypotheses and to separate visual signal from noise. [G1,G2]
\item \textbf{T3 - Match selections to similar examples} based on hidden state activation pattern. A matched phrase should have intuitively similar characteristics as the selection to support or reject a hypothesis. [G2]
\item \textbf{T4 - Align textual annotations} visually to matched phrases. These annotations allow the user to compare the learned representation with alternative structural hypotheses such as part-of-speech tags or known grammars. The set of annotation data should be easily extensible. [G2,G3]
\item \textbf{T5 - Provide a general interface} that can be used with any RNN model and text-like dataset. It should make it easy to generate crowd knowledge and trigger discussions on similarities and differences between a wide variety of models. [G3]
\end{itemize}

\noindent
This list of tasks provided a guideline for the design of \textsc{LSTMVis}. In addition, tasks (T1-T4) define the core interaction mechanisms for discovery with \textsc{LSTMVis}: Visualize (Section~\ref{sec:timeline}) -- Filter \& Select (Section~\ref{sec:selectView}) -- Match \& Align (Section~\ref{sec:matchView}). We will first describe the implementation of these interactions and later demonstrate their application to multiple use cases in Section~\ref{sec:usecases}. 

\section{Design of \textsc{LSTMVis}}
\label{sec:design}

\textsc{LSTMVis} is composed of two major visual analysis components. The \textit{Select View} (Section~\ref{sec:selectView}) supports the formulation of a hypothesis (T2, G1) by using a novel visual encoding for hidden state changes (T1, Section~\ref{sec:timeline}). The \textit{Match View} (Section~\ref{sec:matchView}) allows refinement of a hypothesis (T3, T4, G2) while remaining agnostic to the underlying data or model (T5). 

Our design decisions are the outcome of a long iterative process. First, we developed several interactive, low-fidelity prototypes of varying complexity with our expert users to highlight different aspects of the data (Figure~\ref{fig:design}). After developing a complete system, we released it to a broader audience online to collect long-term feedback (Section~\ref{sec:longterm}). Based on the feedback from this release we developed the current version several months later. 

\subsection{Visualization of Hidden State Changes}
\label{sec:timeline}

Visualizing the progression of hidden state vectors $\mathbf{h}_1, \ldots, \mathbf{h}_T$ along a sequence of words (time-steps) is at the core of \textsc{LSTMVis}.  In the following, we refer to hidden state as one dimension of the $D$-dimensional hidden state vector.

\begin{figure}
\centering
\includegraphics[]{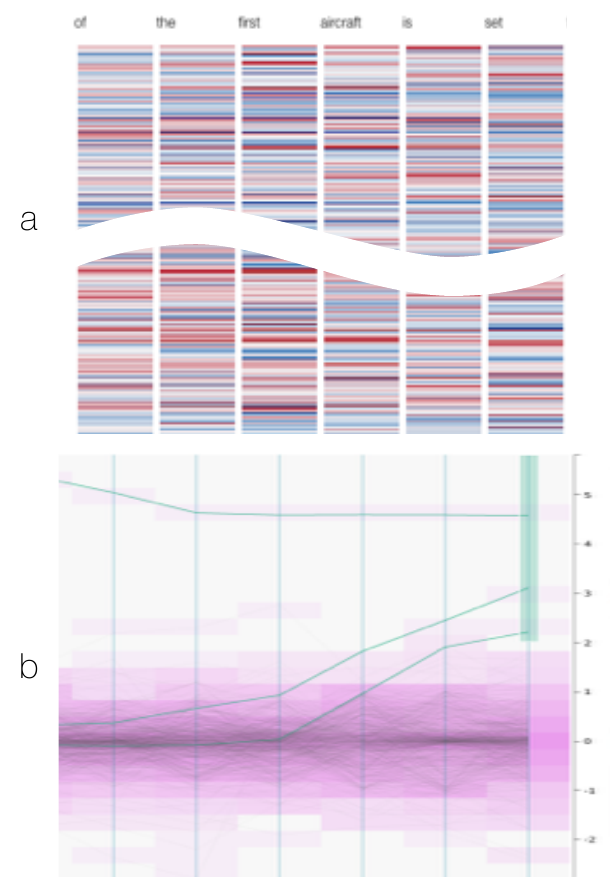}
\caption{Two early-stage prototypes of the system. (a) Hidden state vectors are encoded as heatmaps over time. This style places emphasis on the relationships between neighboring (vertically adjacent) states, which has no particular meaning for this model. (b) A selection prototype utilizing parallel coordinates. This prototype emphasized selections based on small movements of state values directly on the plot, which made it difficult to specify connections  between hidden state values and source text.}
\label{fig:design}
\end{figure}

\begin{figure*}[ht]
  \centering
  \includegraphics[width=\textwidth]{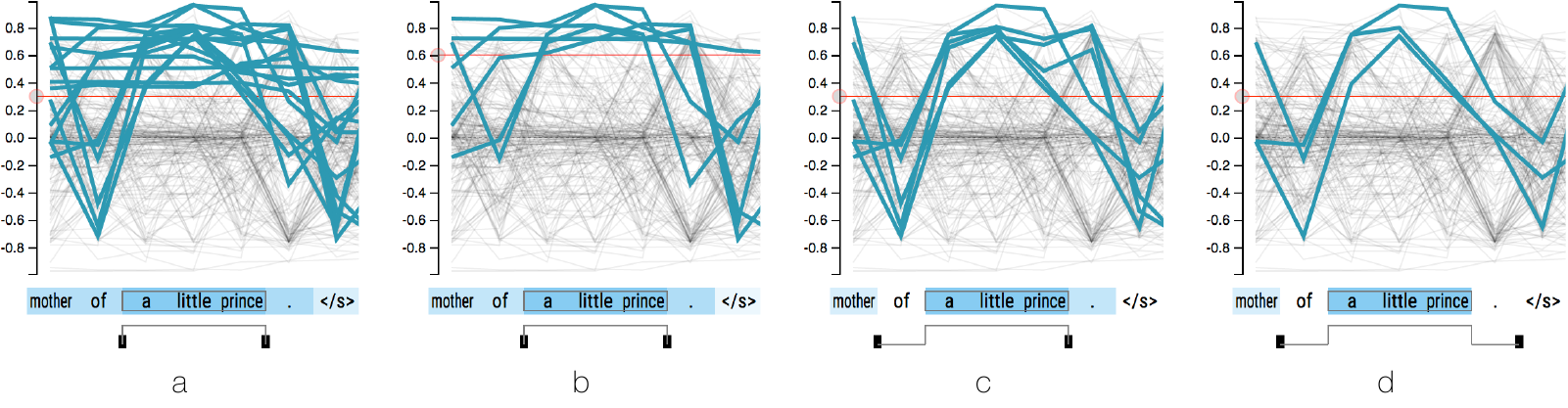}
  
  \caption{The hypothesis selection process. (a) The selection covers \texttt{a little prince} and has a threshold $\ell = 0.3$. Blue highlighted hidden states are selected. (b) The threshold $\ell$ is raised to $0.6$. (c) The pattern plot in the bottom is extended left, eliminating hidden states with values above $\ell$ after reading \texttt{of} (one word to the left). (d) The pattern plot is additionally extended right, removing hidden states above the threshold after reading ``\texttt{.}'' (one word to the right). I.e., only hidden states are selected with the following pattern: below threshold $\ell$ for one word before -- above $\ell$ during \texttt{a little prince} -- below $\ell$ for one word after.}
\label{fig:zeroslider}
\end{figure*}

In the related literature, it is common to encode hidden state vectors as a heatmap along the time-axis (Figure~\ref{fig:design}(a)). This style has been favored as a view of the complete hidden state vectors $\mathbf{h}_1, \ldots, \mathbf{h}_T$.  However, this approach has several drawbacks in an interactive visualization. Foremost, the heatmaps do not scale well with increasing dimensionality $D$ of hidden state vectors. They use a non-effective encoding for the most important information, i.e. hidden state values by color hue. Additionally, they emphasize the order of hidden states in each vector, but this relative order of abstract hidden states is not actually used by the model itself.

We decided to consider each hidden state as a data item and
time-steps as dimensions for each data item in a parallel coordinates plot. Doing so, we encode the hidden state value using the more effective visual variable \textit{position}. Figure~\ref{fig:design}(b) shows the first iteration on using a parallel coordinates plot. The number of data points along the plot is additionally encoded by a heatmap in the
background to emphasize dense (e.g., around the zero value) and sparse regions. 

However, it was very cumbersome to formulate a hypothesis for a longer range by adjusting many y-axis brush selectors at a fine granularity. Allowing the user to perform selection by directly manipulating the hidden state values felt decoupled from the original source of information---the text. The key idea to facilitate this selection process was to allow the
user to easily discretize the data based on a threshold and select on and off ranges directly on top of the words (Section~\ref{sec:matchView}). 

\begin{figure}[htbp]
\centering
\includegraphics[width=.9\columnwidth]{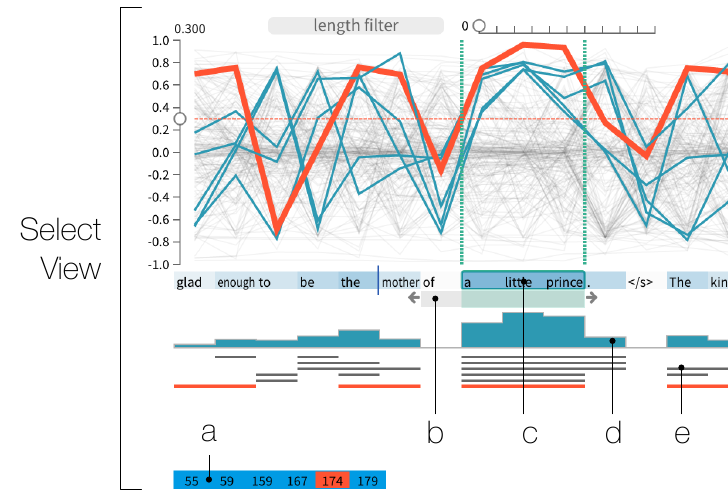}
\caption{Snippet of the first complete prototype. While the list of selected cells (a) and the brushing method (c) remained in the final version, we modified the pattern plot (b) and omitted the redundant encodings (d) and (e).}
\label{fig:V1}
\end{figure}

Figure~\ref{fig:V1} shows the first complete prototype that we put online to collect long-term user feedback (Section~\ref{sec:longterm}). 
Several user comments led us to believe that the redundant encoding of hidden states that are ``on'' in the Select View (Figure~\ref{fig:V1}d,e) was not well understood. In the final design shown on the top in Figure~\ref{fig:teaser} we omitted this redundant encoding for the sake of clarity and to highlight wider regions of text. The x-axis is labeled with the word inputs $w_1, \ldots, w_T$ for the corresponding time-step. If words do not fit into the fixed width for time steps they are distorted. Figure~\ref{fig:teaser} shows the movement of a hidden state vector through an example sequence.

\subsection{Select View}
\label{sec:selectView}

The Select View, shown in the top half of Figure~\ref{fig:teaser}, is centered around the parallel coordinates plot for hidden state dynamics. The full plot can be difficult to comprehend directly. Therefore, \textsc{LSTMVis} allows the user to formulate a hypothesis (G1) about the semantics of a subset of hidden states by selecting a range of words that may express an interesting property. For instance, the user may select a range within a shared nesting levels in tree-structured text (Section~\ref{sec:paren}), a representative noun phrase in a  text corpus (Section~\ref{sec:use_case_noun}), or a chord progression in a musical corpus(Section~\ref{sec:music}). 

To select, the user brushes over a range of words that form the pattern of interest (Figure~\ref{fig:teaser}c). In this process, she implicitly selects hidden states that are ``on'' in the selected range. The dashed red line on the parallel coordinates plot (Figure~\ref{fig:teaser}t) indicates a user-defined threshold value, $\ell$, that partitions the hidden states into ``on'' (timesteps $\geq \ell$) and ``off'' (timesteps $< \ell$) within this range.  

In addition to selecting a range, the user can modify the
pattern plot below (Figure~\ref{fig:teaser}d) to define that hidden states must also be ``off'' immediately before or after the selected range. Figure~\ref{fig:zeroslider} shows different combinations of pattern plot configurations and the
corresponding hidden state selections. Using these selection criteria the user creates a set of selected hidden states $\mathcal{S}_1 \subset \{1 \ldots D\}$ that follow the specified on/off pattern w.r.t. the defined threshold. 

To assist with the selection of ranges, the user can make use of a heatmap underlying the word sequence which depicts how many of the selected hidden states are ``on'' for each word (Figure~\ref{fig:teaser}c). This indicates repetitions of hidden state patterns in the close local neighborhood. 

Additionally, a user can add aligned tracks of textual annotations (meta tracks) to the selection view. E.g., she can visualize part-of-speech (POS) annotations or named entity recognition (NER) results. The feature of meta tracks is the result of feedback from multiple online users asking us to include the tracks. Some users that used the tool for training also wanted to see the top K predictions (outcomes) for each word. Figure~\ref{fig:teaser} shows examples for POS (e1) and topK (e2). 

\textsc{LSTMVis} also provides several convenience methods to navigate to specific time steps. Buttons on the timeline can be used to move forward and backward (Figure~\ref{fig:teaser}-i2). \textsc{LSTMVis} also offers
search functionality to find specific phrases. The selection
panel on the top can be used to efficiently switch between the
different layers (T5). The word-width can be decreased and increased (Figure~\ref{fig:teaser}-i1) to provide support for different average word lengths (e.g., character-based models vs. word-based models). 
The option to turn heatmaps on and off in Match View (Figure~\ref{fig:teaser}g1 and g2) provides convenience when working with many mate tracks.

These interaction methods allow the user to define a hypothesis as word range which results in the selection of a subset of hidden states following a specified pattern w.r.t. a defined threshold (T2, G1) that only relies on the hidden state vectors themselves (T5). To refine or reject the hypothesis the user can then make use of the Match View.

\begin{figure*}[ht]
  \centering
  \includegraphics[]{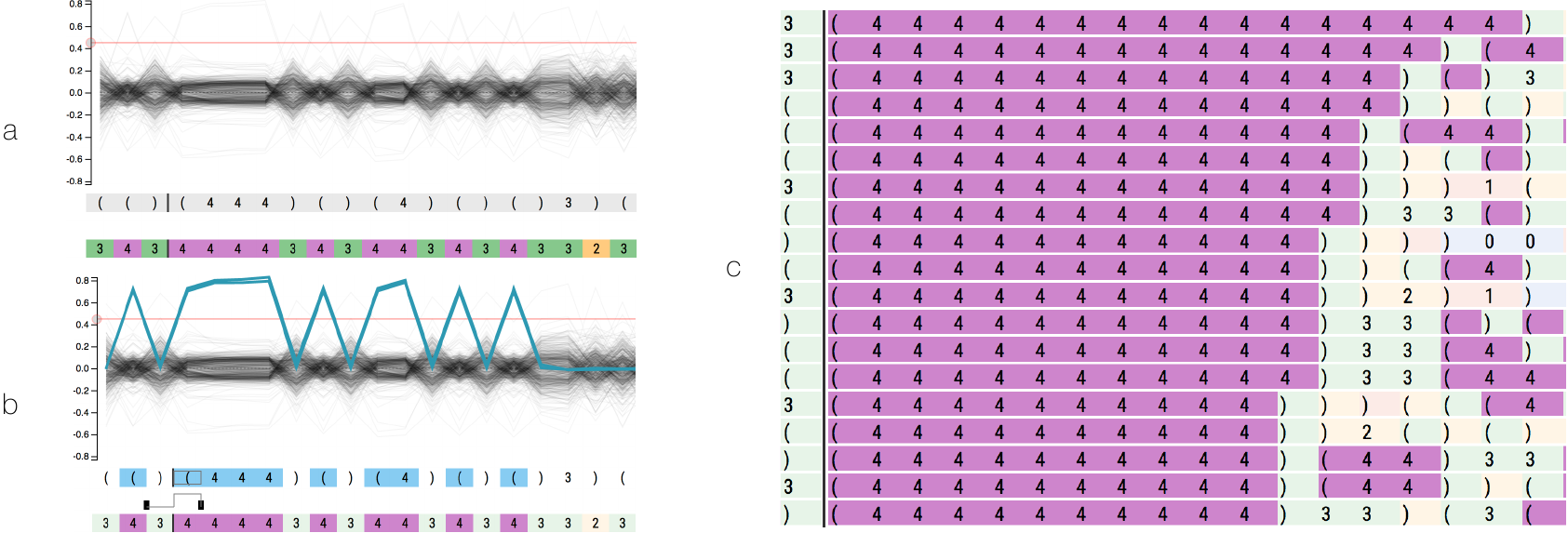} 
  \caption{Plot of a phrase from the parenthesis synthetic language. (a) The full set of hidden states. Note the strong movement of states at parenthesis boundaries. (b) A selection is made at the start of the fourth level of nesting. Even in the select view it is clear that several hidden states represent a four-level nesting count. In both plots the meta-track indicates the nesting level as ground truth. (c) The result of matching the selected states indicates that they seem to capture nesting level 4 phrases of variable length. }
\label{fig:paren}
\end{figure*}

\subsection{Match View}
\label{sec:matchView}

The Match View (Figure~\ref{fig:teaser}b) provides evidence for or against the selected hypothesis. The view provides a set of relevant matched phrases that have similar hidden
state patterns as the phrase selected by the user. 

With the goal of maintaining an intuitive match interface, we define the matches to be ranges in the data set that would have lead to a similar set of on hidden states under the selection criteria (threshold, on/off pattern). Formally, assume that the user has selected a threshold $\ell$ with hidden states $\mathcal{S}_1$ and has not limited the selection to the right or left further, as shown in Figures~\ref{fig:zeroslider}a and \ref{fig:zeroslider}b. We rank all possible candidate ranges in the dataset starting at time $a$ and ending at time $b$ with a two step process

\begin{enumerate}
\item Collect the set of all hidden states that are ``on'' for the range, \[ \mathcal{S}_2 = \{c \in \{1\ldots D\} : h_{t,c} \geq \ell  \text{\ for all \ } a \leq t \leq b \} \] 
\item Rank the candidates by  the number of overlapping states $|\mathcal{S}_1 \cap \mathcal{S}_2|$ using the inverse of the number of additional ``on'' cells $-|\mathcal{S}_1 \cup \mathcal{S}_2|$ and candidate length $b -a$ as tiebreaks.
\end{enumerate}

If the original selection is limited on either side (as, e.g., in Figures~\ref{fig:zeroslider}c or \ref{fig:zeroslider}d), we modify step (2) to take this into account for the candidates. For instance, if there is a limit on the left, we only include state indices $c$ in $\mathcal{S}_2$ in that also satisfy $h_{a-1,c} < \ell$. For efficiency, we do not score at all possible candidate ranges (datasets typically have $T > $ 1 million). We limit the candidate set by filtering to ranges with a minimum number of hidden states from $\mathcal{S}_1$ over the threshold $\ell$. These candidate sets can be computed efficiently using run-length encoding.

From the matching algorithm, we retrieve the top 50 results which are shown, one per row each, in a word matrix (e.g., Figure~\ref{fig:teaser}f) located in the Match View. Each cell in the word matrix is linked to a cell in corresponding heatmaps. These heatmaps encode additional information about each word in the matching results. The always available \textit{match count} heatmap (Figure~\ref{fig:teaser}-g1) encodes the number of overlapping states $|\mathcal{S}_1 \cap \mathcal{S}_2|$ for each timestep.

The user can use additional annotations, similar to meta tracks in the Selection View, as heatmaps (T4). We imagine these annotations act as ground truth data, e.g., part-of-speech tags for a text corpora (Figure~\ref{fig:teaser}-g2), or as further information to help calibrate the hypotheses, e.g., nesting depth of tree-structured text. Figure~\ref{fig:teaser} shows how hovering over ``little'' (mouse pointer) leads to highlights in the \textit{match count} heatmap (g1) indicating seven overlapping states
between match and selection for this position. The POS heatmap (g2) indicates that the word at this position acts as adjective (ADJ).

The heatmap colors can be mapped directly to the background of the matches (Figure~\ref{fig:teaser}h) as a simple method to reveal pattern across results (Figure~\ref{fig:teaser}f). 
Based on human identifiable patterns or alignments, the matching and mapping methods can lead to further data analysis or a refinement of the current hypothesis.

\subsection{Design Considerations and Limitations}
 LSTMVis operates around as a bottom-up approach starting from a seed hypothesis and searching for similarities across the whole dataset. During the project, we experimented with top-down approaches following the Shneiderman mantra (overview first -- zoom and filter later). We found that global overviews, such as projections of hidden state values or summary statistics, did little to reveal interpretable patterns on large datasets. 
LSTMVis differs from related work, in that instead of working with the prediction output, it operates directly on the hidden states to identify relationships. This design decision addresses our target group of architects and trainers who are familiar with model internals. 

Another goal of our tool is to use a representation that is invariant to the ordering of hidden states in the hidden state vector while remaining scalable w.r.t. size of the vector. As addressed in Section ~\ref{sec:timeline}, 
those two conditions are fulfilled by plotting hidden states individually onto parallel coordinates. However, even with this compact representation, applications with $>500$ hidden states become challenging to analyze as a whole. Approaches to address the visual noise are to either filter the hidden states using our interactive methods or to use algorithmic preprocessing, like dimensionality reduction techniques or pruning methods. After testing the latter automated filtering approaches, we decided against using them within the application. We found that if the tool loses its inherent connection to the data, results were less interpretable to the user.

\subsection{Technical Design and Implementation}
\label{sec:implementation}

\textsc{LSTMVis} consists of two modules, the visualization system and the RNN modeling component. The source code with documentation and some example models are available at \url{lstm.seas.harvard.edu}.

The visualization is a client-server system that uses Javascript and D3 on client side and Python, Flask, Connexion, h5py, and numpy on server side. Timeseries data (RNN hidden states and input) is loaded dynamically through HDF5 files. Optional annotation files can be specified to map categorical data to labels (T4). New data sets can be added easily by a declarative YAML configuration file.

The RNN modeling system is completely separated from the visualization to allow compatibility with any deep learning framework (T5). For our experiments we use the Torch framework. We trained our models separately and exported results to the visualization.

\section{Use Cases}
\label{sec:usecases}

In experimenting with the system we trained and explored many different RNN models, datasets and tasks, including word and character language models, neural machine translation systems, auto-encoders, summarization systems, and classifiers. Additionally, we also experimented with other types of real and synthetic input data. 

In this section we highlight three use cases that demonstrate the general applicability of \textsc{LSTMVis} for the analysis of hidden states.

\subsection{Proof-of-Concept: Parenthesis Language}
\label{sec:paren}

As proof of concept we trained an LSTM as language model on synthetic data generated from a very simple counting language with a parenthesis and letter alphabet $\Sigma= \{$\verb|( ) 0 1 2 3 4 |$\}$. The language is constrained to match parentheses, and nesting is limited to at most 4 levels deep. Each opening parenthesis increases and each closing parenthesis decreases the nesting level, respectively. Numbers are generated randomly, but are constrained to indicate the nesting level at their position. For example, a string in the language might look like:

\hfill\includegraphics[width=.8\columnwidth]{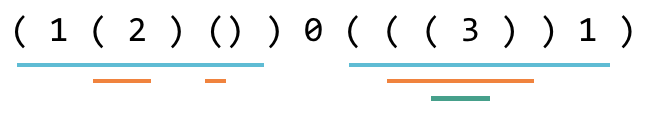}\hspace*{\fill}

\noindent
Blue lines indicates ranges of nesting level $\geq$1, orange lines indicate nesting level $\geq$2, and green lines indicate nesting level $\geq$3.

To analyze this language, we view the states in \textsc{LSTMVis}. An example of the cell states of a multi-layer 2x300 LSTM modelis is shown in Figure~\ref{fig:paren}(a). Here even the initial parallel coordinates plot shows a strong regularity, as hidden state changes occur predominately at parentheses.

Our hypothesis is that the hidden states mainly reflex the nesting level. To test this, we select a range spanning nesting level four by selecting the phrase \verb| ( 4|. We immediately see that several hidden states seem to cover this pattern and that in the local neighborhood several other occurrences of our hypothesis are covered as well, e.g., the empty parenthesis and the full sequence \verb| ( 4 4 4 | . This observation nicely confirms earlier work that
demonstrates simple context-free models in RNNs and LSTMs~\cite{wiles1998recurrent,gers2001lstm}.

\subsection{Phrase Separation in Language Modeling}
\label{sec:use_case_noun}

\begin{figure}[ht]
  \centering
  \includegraphics[]{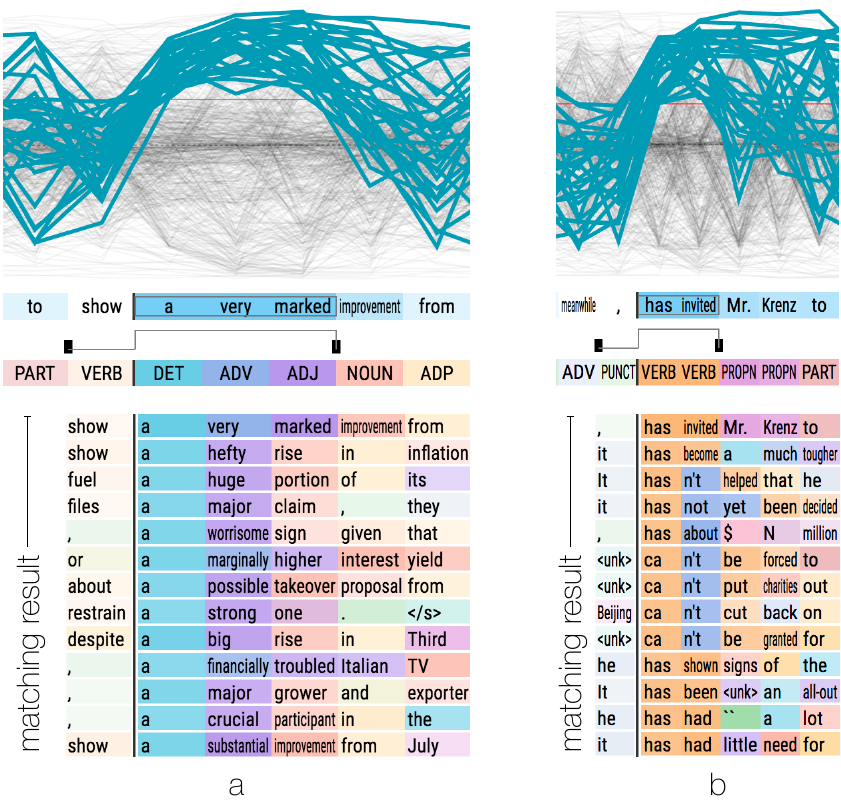}
  \label{fig:hyp}
  \caption{Phrase selections and match annotations in the Wall Street Journal. (a) The user selects the phrase \texttt{a very marked improvement} (turning off when improvement is seen). The matches found are entirely other noun phrases, and start with different words. Note that here ground-truth noun phrases are indicated with a sequence of colors: cyan (DET), blue (ADV), violet (ADJ), red (NOUN). (b) We select a range starting with \texttt{has invited}. The results are various open verb phrases as sequence of colors orange (VERB) and blue (ADV). Note that for both examples the model can return matches of varying lengths.
  }
\end{figure}

Next we consider the case of a real-world natural language model from the perspective of an architect interested in the structure of the internal states and how they relate to underlying properties. For this experiment we trained a 2-layer LSTM language model with 650 hidden states on the Penn Treebank~\cite{marcus1993building} following the medium-sized model of \cite{Zaremba2014}. While the model is trained for language modeling (predict the next word), we were interested in seeing if it additionally learned properties about the underlying language structure. To test this, we additionally include linguistic annotations in the visual analysis from the Penn Treebank. We experimented with including part-of-speech tags, named entities, and parse structure. 

Here we focus on the case of \textit{phrase chunking}. We annotated the dataset with the gold-standard phrase chunks provided by the CoNLL 2003 shared task~\cite{tjong2003introduction} for a subset of the
treebank (Sections 15-18). These include annotations for noun phrases and verb phrases, along with prepositions and several other less common phrase types.

While running experimental analysis, we found a strong pattern that selecting noun phrases as hypotheses leads to almost entirely noun phrase matches.  Additionally, we found that selecting verb phrase prefixes would lead to primarily verb phrase matches. In Figure~\ref{fig:hyp} we show two examples of these selections and matches.

\begin{figure}[ht]
  \centering
  \includegraphics[width=\linewidth]{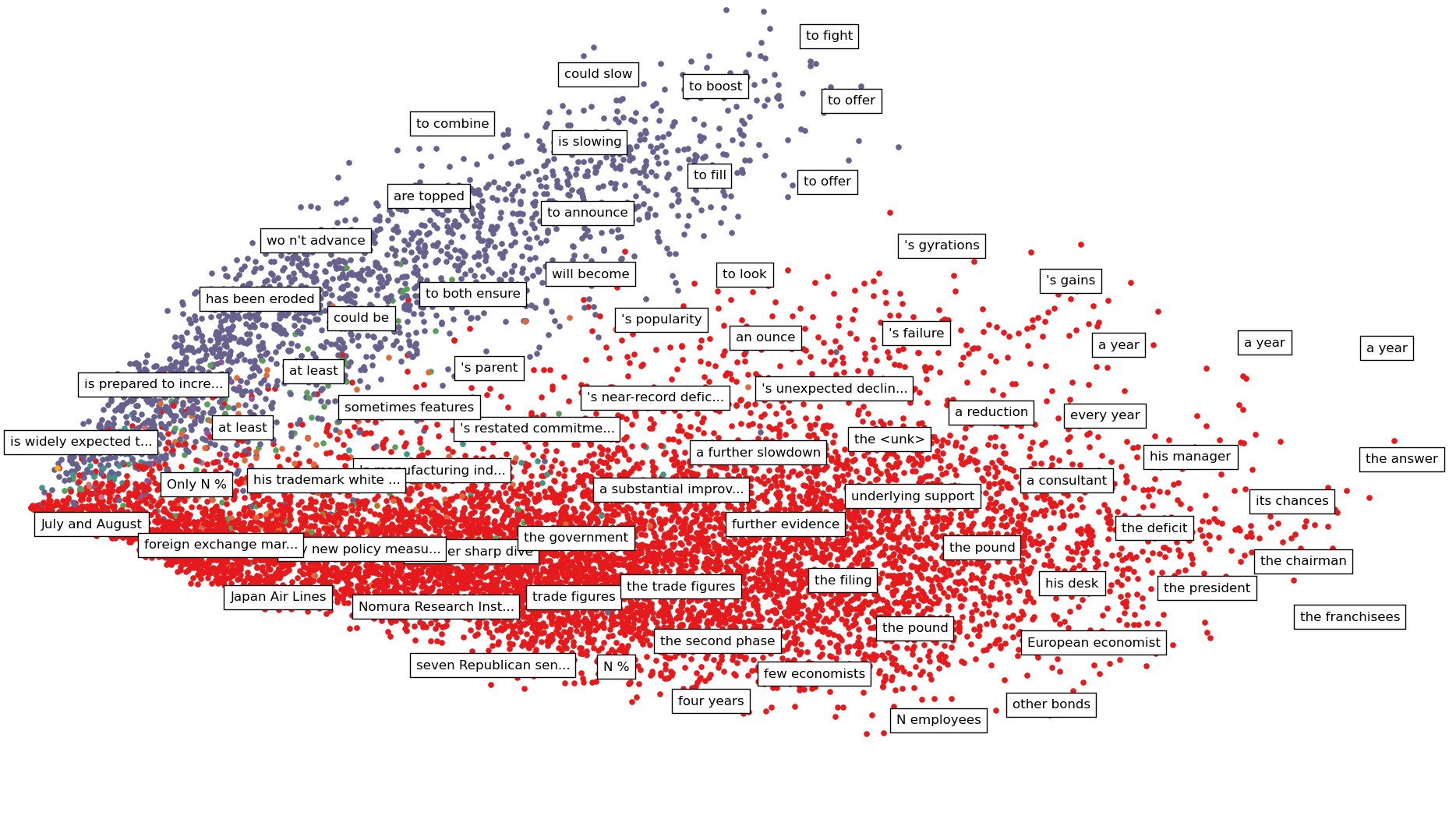}
  \label{fig:pca}
  \caption{PCA projection of the hidden state patterns ($\mathcal{S}_1$) of all multi-word phrasal chunks in the Penn Treebank, as numerical follow-up to the phrase chunking hypothesis. Red points indicate noun phrases, blue points indicate verb phrases, other colors indicate remaining phrase types. While trained for language modeling, the model separates out these two phrase classes in its hidden states. }
\end{figure}

The visualization hints that the model has implicitly learned a representation for language modeling that can differentiate between the two types of phrases. Of course the tool itself cannot confirm or deny this type of hypothesis, but the aim is to provide clues for further analysis. We can check, outside of the tool, if the model is clearly differentiating between the classes in the phrase dataset. To do this we compute the set $\mathcal{S}_1$ for every noun and verb phrase in the shared task. We then run PCA on the vector representation for each set. The results are shown in Figure~\ref{fig:pca}, which shows that indeed these on-off patterns are enough to partition the noun phrases and verb phrases.

\subsection{Biological Sequence Analysis}
\label{sec:usecase_bio}

RNN models are now commonly used for time-series analysis 
outside of the space of text processing. One area of interest
is in biological sequence analysis for genomics where deep neural networks 
are directly applied to sequences of DNA for tasks such as classification
and sequence labeling. We consider models trained 
for regulatory marker prediction, where neural networks are used to predict 
binding sites aligned to sequences of the genome, a task that has been 
recently explored with both CNN \cite{deepsea} and RNN models \cite{danq}.
For this case study, we collaborated with a domain expert \textit{trainer}
who approached us after becoming aware of the release of our tool. He had employed a mixed RNN/CNN model that was trained over a genomic dataset made up of a 2.3 billion base pair long nucleotide sequence for this problem of 
regulatory marker prediction. 

\begin{figure}[ht]
\centering
\includegraphics[width=0.8\linewidth]{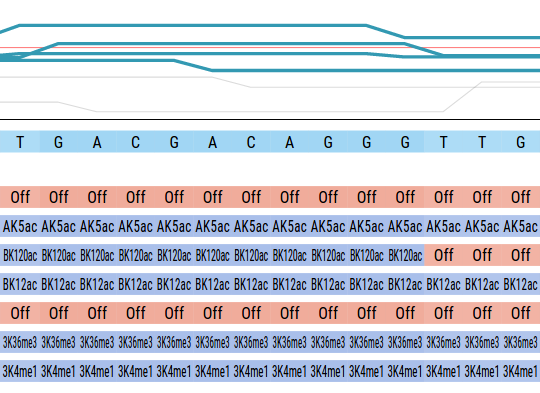}
\label{fig:bsa}
\caption{Biological Sequence Analysis. A selected region of the genome with 
	seven different aligned annotation tracks. Each track indicates the 
    0/1 prediction of protein binding sites at each time step. Tracks can be 
    activated/deactivated as part of the debugging process to compare predictions,
    ground truth, input properties, and other user-specified annotations.}
\end{figure}

Notably this problem differs in several crucial ways
from our previous use case: the granularity of the input is much smaller (base pairs as opposed to words), the training objective is different (0/1 binding site labels as opposed to the next word), and the problem is known to exhibit much longer term dependencies due to its latent 3D structure.  Conversations around these issues led us to introduce several iterative features to our tool. These include the (a) ability to zoom in and out of the timeseries representation to adjust for granularity, (b) use of auto-scaling axes ranges to allow for switching between data of different activation ranges, such as CNN outputs, and most notably (c) ability to add/remove arbitrarily many annotation label along the word sequence timeline, to allow domain experts to view both the predictions and ground truth as an annotation track associated with the features at each time step. Figure~\ref{fig:bsa} shows an example of the last feature which shows the predict binding locations for several different proteins along a region of DNA. 

For this application the tool has so far been used primarily as a debugging assistant, providing a way for an expert in an ongoing research project to analyze the output of a 
trained model and make adjustments based on mistakes of the model. For instance, in early use of the tool, the researcher
noticed that one layer of the network was using very few of the available states, and another 
had significant artifacts from a poorly aligned convolutional layer. These observations have helped provide feedback for subsequent experimental design. Furthermore, recent work indicates an increasing interest in LSTMVis in the biomedical domain and for genomics (see Section~\ref{sec:longterm}).

\subsection{Musical Chord Progressions}
\label{sec:music}

Past work on LSTM structure has emphasized cases where single hidden states are semantically interpretable. For text and biological data sets, we found that with a few exceptions (quotes, brackets, and commas) this was rarely the case. However, for datasets with more regular long-term structure, single states could be quite meaningful. 

As a simple example, we collected a large set of songs with annotated chords for rock and pop songs to use as a training data set, 219k chords in total. We then trained an LSTM language model to predict the next chord $w_{t+1}$ in the sequence, conditioned on previous chord symbols (chords are left in their raw format). 

\begin{figure}
  \centering
  \includegraphics[width=\linewidth]{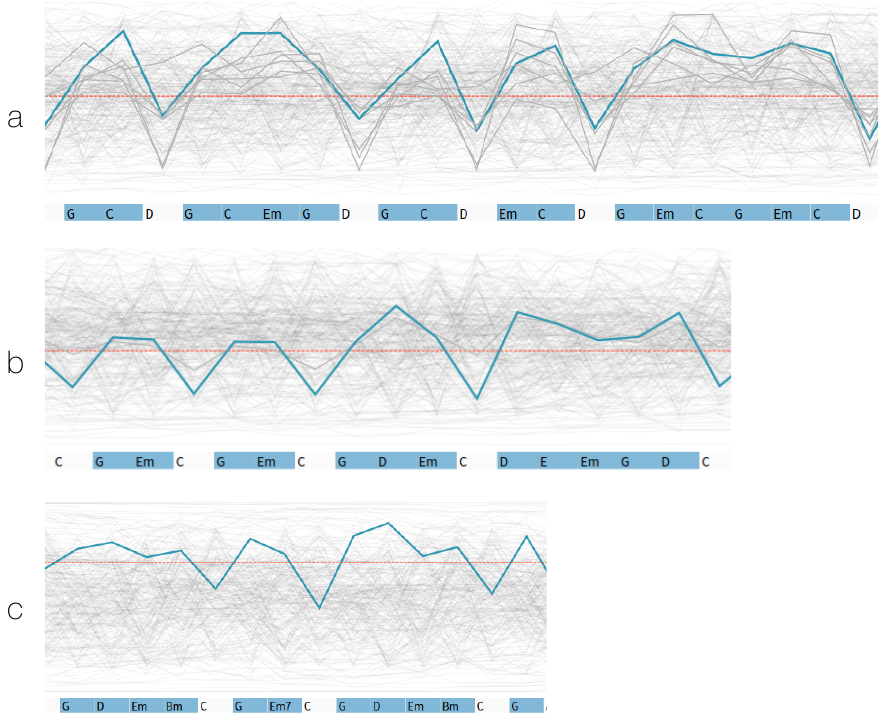}
  \label{fig:progression}
  \caption{Three examples of single state patterns in the guitar chord dataset. (a) We see several permutation of the very common I - V - vi - IV progression (informally, the ``Don't Stop Believing'' progression). (b) We see several patterns ending in a variant of the I- vi- IV- V (the 50's progression). (c) We see two variants of I - V - vi -iii - IV - I (beginning of the Pachelbel's Canon progression). Chord progression patterns are based on \url{http://openmusictheory.com/}.
}
\end{figure}

When we viewed the results in \textsc{LSTMVis} we found that the regular repeating structure of the chord progressions is strongly reflected in the hidden states. Certain states will turn on at the beginning of a standard progression, and remain on though variant-length patterns until a resolution is reached. In Figure~\ref{fig:progression}, we examine three very common general chord progressions in rock and pop music. We select a prototypical instance of the progression and show a single state that captures the pattern, i.e., remains on when the progression begins and turns off upon resolution.

\section{Long-Term Case Study}
\label{sec:longterm}

Shneiderman and Plaisant˜\cite{Shneiderman2006Strategies} propose strategies for multi-dimensional in-depth long-term case (MILC) studies to evaluate information visualization tools. They observe that ``scaling up by an order of magnitude in terms of number of users and duration of the observations is clearly beneficial.'' We decided to adopt their core ideas and report on qualitative feedback and quantitative success indicators after the open-source release of \textsc{LSTMVis} in June 2016.
  
We created a webpage and a video that introduces the core ideas of \textsc{LSTMVis} at \url{lstm.seas.harvard.edu}. The webpage provides an opportunity for users to leave comments. To advertise the tool online we followed a social media strategy that included collecting followers on Twitter, using broadcast media such as Reddit and Hackernews, and inviting editors of popular machine learning blogs to write guest articles.

The webpage contains a demo system with example datasets that allows us to acquire a large set of logging data. We noticed that users often try to reproduce the scenarios that are explained in the online video. To allow users to share insights from their exploratory analysis, we ensured that our URL parameter string contains all necessary information to replay and share a scenario. 

We collected logging information about our webpage using Google Analytics. Within the first 7 days, the web page received $\sim$5,600 unique users, with 49\% of traffic coming  through social media and 39\% arriving directly. The social media traffic was dominated by channels that we used to advertise the tool (Twitter 40\%, Reddit 26\%). To our surprise we also observed substantial traffic from channels that we did not contact directly (e.g., Sina Weibo 11\%, Google+ 9\%). After 300 days we recorded $\sim$19,500 page views with shrinking user traffic from social media (31\%). At that point most users used search (22\%) or accessed the  webpage directly (39\%). Only a small percentage (10\%) of users tried the online demo. Most of these users used the datasets shown in the explanation video and did not explore further.

Our open source code release was stable yet simple enough to ``...ensure that the tool has a reasonable level of reliability and support for basic features''~\cite{Shneiderman2006Strategies}. For easy adoption, we provide user documentation that explains what data can be used and how to use the tool. We also provide convenience tools to prepare the data for import (Section~\ref{sec:implementation}). We asked students to act as testers for our source code. Based on their feedback we made several improvements to the installation process. For example, our prototype required NodeJS to resolve client-side dependencies. By providing the required libraries within our repository we removed the cumbersome step of installing NodeJS for our target audience.  We observe that around 400 programmers liked the project (stars on GitHub) and over 95 practitioners copied (forked) the project to make custom modifications. We think that this demonstrate a reasonable interest in the system after only 300 days considering its highly specialized target audience. 

Furthermore, we observe adoption of the tool for several documented use cases. Evermann et.al.~\cite{Evermann2017} describe the application of LSTMVis to understand a trained model for a business process intelligence dataset. Liu et.al.~\cite{liu2016modeling} use LSTMVis in experiments to investigate language variants and vagueness in website privacy policies. We see an increasing interest to apply our tool for biomedical and genomic data. This is indicated by the use case we described in Section~\ref{sec:usecase_bio} or, e.g., in Ching et.al.\cite{Ching142760}

Besides the quantitative observations we also collected qualitative feedback from comments on the webpage, GitHub tickets (feature requests), and in-person presentations of the prototype. This qualitative feedback led us to make several changes to our system that were discussed in Section~\ref{sec:design}. 

In retrospective, conducting a long-term user study benefited the project at multiple stages. Preparing the release of the prototype required us to focus strongly on simplicity, usability, and robustness of our tool. This lead to many small improvements to an internal prototype. The design iterations we inferred from user feedback strengthen the tool further. We think, that planning a successful long-term study requires four core ingredients: (1) reach out to your target audience by describing your approach (webpage, video) and inform them using social media, email, etc. (2) allow users to play with your tool by setting up a demo server, (3) allow engagement and experimentation with your tool by providing sufficiently documented, easily adoptable source code, and (4) make it as simple as possible for users to give you feedback via discussion forums, reported issues, or in person. 
During the study, we found it to be crucial to continuously engage with users and quickly take action based on their feedback.

\section{Conclusion}
\label{sec:conclusion}

\textsc{LSTMVis} provides an interactive visualization to
facilitate data analysis of RNN hidden states. The tool is based on direct inference where a user \textit{selects} a range of text to represent a hypothesis and the tool then \textit{matches} this selection to other examples in the data set. The tool easily allows for external annotations to verify or reject hypothesizes. It only requires a time-series of hidden states, which makes it easy to adopt for a wide range of visual analyses of different data sets and models, and even different tasks (language modeling, translation etc.). 

To demonstrate the use of the model we presented several case studies of applying the tool to different data sets. Releasing the tool and source code online allows us to collect long-term  user feedback that has already led us to make several improvements. 
In future work, we will explore how the wide variety of application cases can be adopted -- beyond our imagined use cases and user groups. 
As example for such a use case, we got contacted by a highschool student using LSTMVis to learn about RNN methods. While we did not optimize
for a learning scenario, we are now thinking about datasets and a blog publication that focus on learning. Another interesting 
future work is to support the user role of \textit{end users} with simplified views on internals of RNN to help explaining specific model behavior.

\acknowledgments{This work was supported in part by the Air Force Research Laboratory, DARPA grant FA8750-12-C-0300, Oracle Research grant, and NIH grant U01CA198935.}

\bibliographystyle{abbrv-doi}
\bibliography{template}
\end{document}